\documentclass[letterpaper, 10 pt, conference]{ieeeconf}  
\usepackage{amssymb}
\usepackage{amsmath}
\usepackage{amsmath,mathtools}
\usepackage{times}
\usepackage{caption}
\usepackage[labelformat=simple]{subcaption}
\usepackage{graphicx}
\usepackage{graphics}
\usepackage{blkarray}
\usepackage{booktabs}
\usepackage{xcolor} 
\usepackage{bbold}
\usepackage{caption}
\usepackage{subcaption}
\usepackage{multirow}
\usepackage{comment}
\usepackage{epstopdf}
\usepackage{tikz}
\usepackage{todonotes}
\graphicspath{{./figures/}}
\usepackage{csquotes}
\usepackage{siunitx}
\usepackage{cite}
\usepackage{epigraph}
\usepackage{empheq}
\usepackage{comment}
\usepackage{cancel}

\usepackage[linesnumbered,ruled,vlined,algo2e]{algorithm2e}

\newcommand{\bt}{\mathcal{T}}

\SetKwRepeat{Do}{do}{while}

\makeatother

\DeclareCaptionLabelSeparator{periodspace}{.\quad}
\IEEEoverridecommandlockouts                              
\newtheorem{remark}{Remark}

\newtheorem{example}{Example}

\newtheorem{problem}{Problem}

\makeatletter
\newcommand*{\rom}[1]{\expandafter\@slowromancap\romannumeral #1@}
\makeatother

\def\eqalignno#1{\let\\=\cr\displ@y \tabskip\@centering
  \halign to\displaywidth{\hfil$\@lign\displaystyle{##}$\tabskip\z@skip
    &$\@lign\displaystyle{{}##}$\hfil\tabskip\@centering
    &\llap{$\@lign##$}\tabskip\z@skip\crcr
    #1\crcr}}
\def\leqalignno#1{\let\\=\cr\displ@y \tabskip\@centering
  \halign to\displaywidth{\hfil$\@lign\displaystyle{##}$\tabskip\z@skip
    &$\@lign\displaystyle{{}##}$\hfil\tabskip\@centering
    &\kern-\displaywidth\rlap{$\@lign##$}\tabskip\displaywidth\crcr
    #1\crcr}}
\begin{document}

\thispagestyle{empty}
\twocolumn
\title{\LARGE \bf
Towards Blended Reactive Planning and Acting using Behavior Trees}

\author{Michele Colledanchise\textsuperscript{1}\thanks{ \textsuperscript{1} iCub Facility, Istituto Italiano di Tecnologia - IIT, Genoa, Italy}, Diogo Almeida\textsuperscript{2} \thanks{\textsuperscript{2}
KTH - Royal Institute of Technology, Stockholm, Sweden.}, and Petter \"Ogren\textsuperscript{2} 
}

\maketitle
\thispagestyle{empty}
\pagestyle{empty}

\begin{abstract}
In this paper, we show how a planning algorithm can be used to automatically create and update a Behavior Tree (BT), 
controlling a robot in a dynamic environment.
The planning part of the algorithm is based on the  idea of back chaining.
Starting from  a goal condition we iteratively select actions to achieve that goal, and if those action have unmet preconditions, they are extended with actions to achieve them in the same way. 
The fact that BTs are inherently modular and reactive makes the proposed solution
blend acting and planning in a way that enables the robot to efficiently
react to external disturbances.
If an external agent undoes an action the robot re-executes it without re-planning,
and if an external agent helps the robot, it skips the corresponding actions, again without re-planning.
We illustrate our approach in two different robotics scenarios.


\end{abstract}
\section{Introduction}
\label{sec:introduction}

Behavior Trees (BTs) were developed within the computer gaming industry as a modular and flexible alternative to Finite State Machines (FSMs). Their recursive structure and usability have made them very popular in industry, which in turn has created a growing amount of attention in academia \cite{colledanchise2017behavior,Bagnell2012b,klockner2013,Colledanchise14iros,hu2015ablation,guerin2015manufacturing}. However, the vast majority of BTs are still manually designed. In this paper, we show how to
automatically create  a BT using a planning algorithm.
The resulting approach allows us to blend planning and acting in a reactive and modular fashion.

\begin{figure}[t!]
    \centering
    \begin{subfigure}[t]{0.485\columnwidth}
        \centering
\includegraphics[width = \columnwidth, ,trim={7cm 2.5cm 8cm 2.5cm},clip]{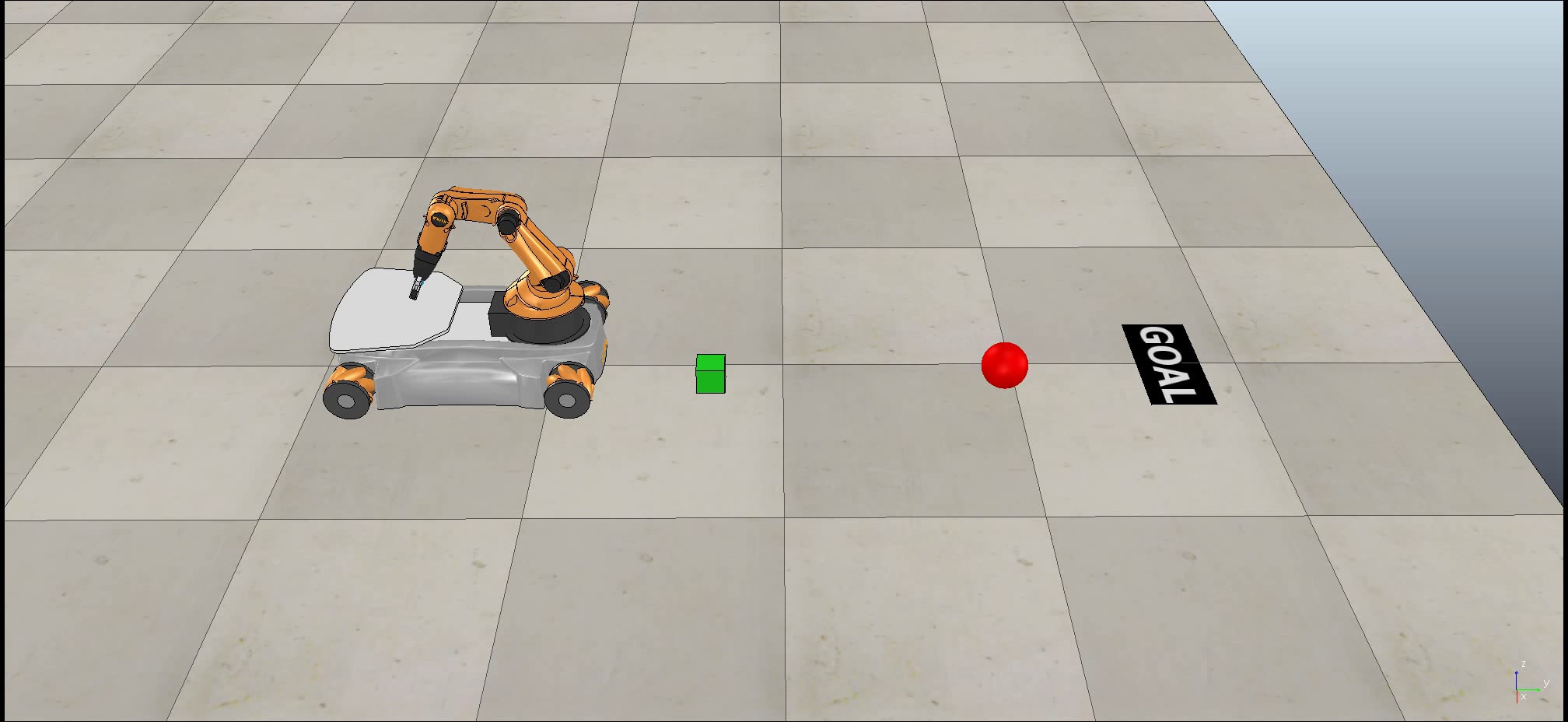}
         \caption{}
    \end{subfigure}%
    ~ 
    \begin{subfigure}[t]{0.485\columnwidth}
        \centering
\includegraphics[width = \columnwidth, trim={7cm 2.5cm 8cm 2.5cm},clip]{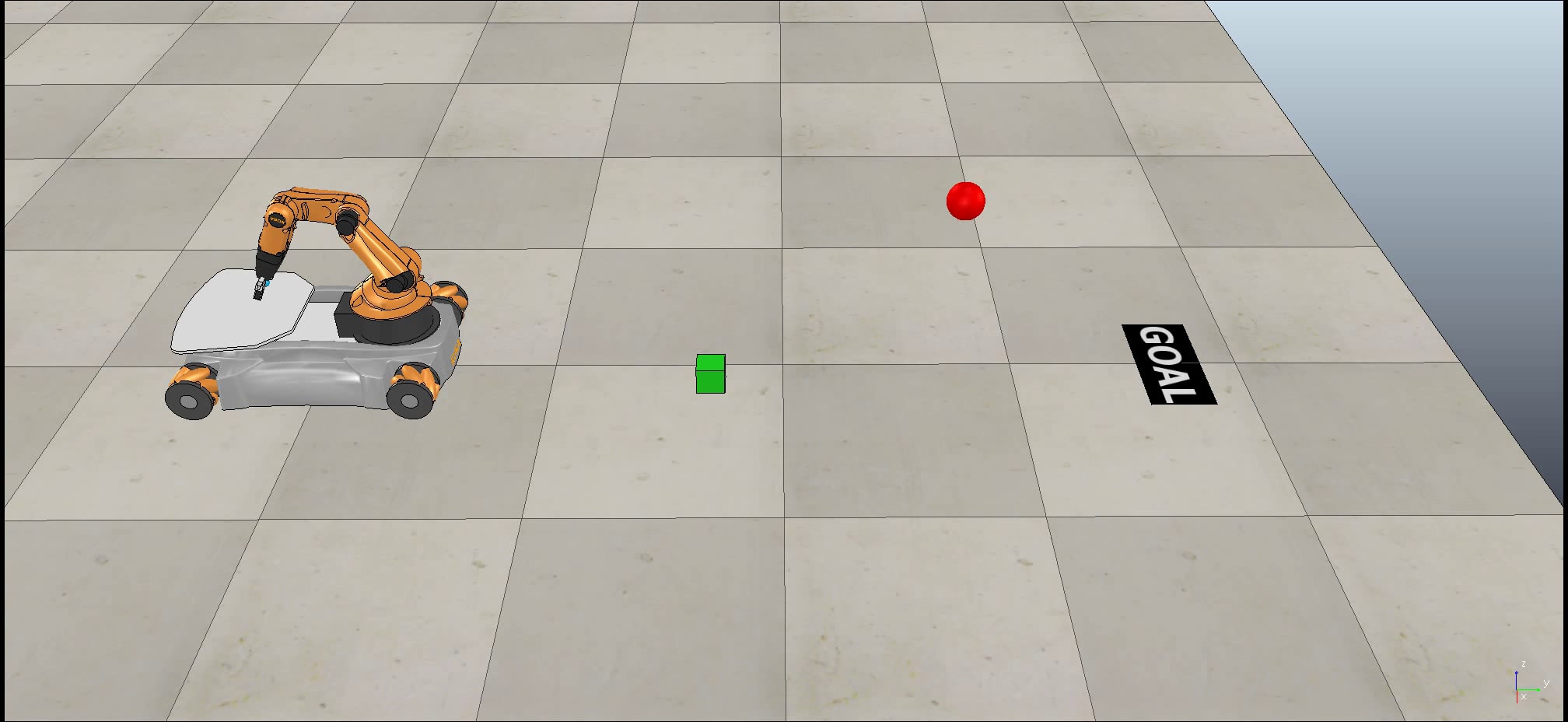}
   \caption{}
    \end{subfigure}
    \vspace{1em}
    \caption{
    A simple example scenario where the
    goal is to place the green cube $C$ onto the goal region $G$. But the fact that the sphere $S$ intermittently  blocks the path must be handled.
    In (a) the nominal plan is \emph{MoveTo(C)$\to$Pick(C)$\to$MoveTo(G)$\to$Drop()} when the sphere suddenly blocks the path. 
     After replanning, the plan is  \emph{MoveTo(S)$\to$Push(S)$\to$MoveTo(C)} 
        \emph{$\to$Pick(C)$\to$MoveTo(G)$\to$Drop()}.
        In (b),  an external agent moves the sphere before being pushed by the agent. Thus the actions concerning the sphere $S$ should be ignored. }
    \label{IN.fig.front}
\end{figure}

To illustrate how the proposed approach blends planning and acting, we use a simple example, depicted in Figure~\ref{IN.fig.front}.
A robot has to plan and execute the actions needed to pick up an object, and place it in a given location. The environment is however dynamic and unpredictable. After pickup, the object might slip out of the robot gripper, or, as shown in Figure~\ref{IN.fig.front}(a), external objects might move and block the path to the goal location, and then move away again, forcing the robot to react once more, see Figure~\ref{IN.fig.front}(b). 
The BT includes reactivity, in the sense that if the object slips out of the robot gripper, it will automatically stop and pick it up again without the need to replan or change the BT.
The BT also supports iterative plan refinement, in the sense that if an object moves to block the path, the original BT is extended to include a removal of the blocking obstacle. Then, if the obstacle is removed by an external actor, the BT reactively skips the obstacle removal, and goes on to pick up the main object without having to change the BT. 

Within the AI community, there has been an increased interest in the combination of planning and acting, \cite{Ghallab14,ghallab2016automated}. In particular, \cite{Ghallab14}  describes two key open challenges, summarized in the following quotes:
\begin{itemize}
\item Challenge 1: ``Hierarchically organized deliberation. This principle goes beyond existing hierarchical planning techniques; its requirements and scope are significantly different. The actor performs its deliberation online"
\item Challenge 2: ``Continual planning and deliberation. The actor monitors, refines, extends, updates, changes and repairs its plans throughout the acting process, using both descriptive and operational models of actions."
\end{itemize}
Similarly, the recent book \cite{ghallab2016automated} describes the need for an agent that 
``reacts to events and extends, updates, and repairs its plan on the basis of its perception".
Finally, the authors of  \cite{ghallab2016automated} also note that most of the current work in action planning yields a static plan, i.e., a sequence of actions that brings the system from the initial state to the goal state. Its execution is usually represented as a classical FSM. However, due to  external agents creating changes in the environment, the outcome of an action can be unexpected. This may lead to situations where the agent replans from scratch on a regular basis, which can be expensive in terms of both time and computational load.

BTs are a graphical mathematical model for reactive fault tolerant task executions. 
They were first introduced in the computer gaming industry \cite{isla2005handling} to control in game opponents, and is now an established tool appearing in textbooks \cite{millington2009artificial,rabin2014gameAiPro,BTBook} and generic game-coding software such as Pygame\footnote{http://www.pygame.org/project-owyl-1004-.html}, Craft AI \footnote{http://www.craft.ai/}, and the Unreal Engine\footnote{https://docs.unrealengine.com/latest/INT/Engine/AI/BehaviorTrees/}. 
BTs are appreciated for being highly modular, flexible and reusable, and have also been shown to generalize other
successful control architectures such as the 
 Subsumption architecture,~\cite{colledanchise2017behavior} and the Teleo-reactive Paradigm~\cite{Colledanchise16iros}.
 So far, BTs are either created by human experts  \cite{ogren,klockner2013,Colledanchise14iros, hu2015ablation, guerin2015manufacturing,klockner2015behavior} or automatically designed using machine learning techniques \cite{pena2012learning,perez2011evolving,colledanchise2015learning} maximizing some heuristic objective function. 

In this paper we propose an automated planning approach to synthesize a BT. The construction of the tree is based on the idea
of backchaining. Starting from the goal condition we find actions that meet those conditions. We then look at the preconditions of 
those actions and try to find actions that satisfy them, and so on.
This is a well known approach, but the novelty lies in the combination with BTs,
exploiting their advantages in terms of \emph{reactivity} and \emph{modularity}, \cite{colledanchise2017behavior},
as compared to e.g., Finite State Machines.

Looking back at the example above, the \emph{reactivity} of BTs enable the robot to pick up a dropped object without 
having to replan at all. The \emph{modularity} enables extending the plan by adding actions for handling the blocking sphere, without having to
replan the whole task. Finally, when the sphere moves away, once again the \emph{reactivity} enables the correct execution without
changing the plan.


%
%

The main contribution of this paper is thus that we show how to iteratively create and refine BTs using a planning algorithm,
and that the result is both reactive and modular, as described above. To the best of our knowledge,
this has not been done before.

The rest of this paper is organized as follows. In Section~\ref{sec:related_work} we present related work,
then in Section~\ref{sec:background} we describe BTs. 
In Section~\ref{sec:problem} we describe the problem that we want to solve,
and in Section~\ref{sec:proposed_approach}  we describe the proposed solution. 
Some simulations are performed in Section~\ref{sec:simulations} to illustrate the approach, before concluding in Section~\ref{sec:conclusions}.

\section{Related work}
\label{sec:related_work}

In this section we briefly summarize related work and compare it with the proposed approach. We focus on Automated Planning   as little work in the literature addresses our objective of automatically generating a BT. 

The planning community has developed solid solutions for solving path-finding problems in large state spaces. Such solution have found successful applications in a wide variety of problems. Nonetheless, numerous planning problems remain  open challenges \cite{Ghallab14, jimenez2012review, Nau15Challenges, latombe2012robot}. 
For example, it was noted by Kaelbling et al. \cite{Kaelbling13}, that there is no systematic planning framework that can address an abstract goal such as "wash dishes" and reason on a long sequence of actions in dynamic or finite horizon environments.  

In the robotic community most of the work focus, without loss of generality, on manipulation planning, where the objective is to have a robot operate in an environment and change the configuration of that environment by e.g., moving objects and opening doors.

Early approaches treated the configuration space as continuous for both object and robot  but used discrete actions \cite{lozano1981automatic,lozano1987handey,latombe2012robot}.

Later work \cite{hauser2011randomized} proposed  so-called \emph{multi-modal planning}, as a generalization of previous approaches,  using different operational \emph{modes} representing different constraint subspaces of the state space. These plans were characterized by  switching between operating a single mode and choosing the mode. Multi-modal planning was then extended to address more complex problems combining a bidirectional search with an hierarchical strategy to determine the operational model of the actions \cite{barry2013hierarchical}. In contrast with our approach,  these works assume a static environment and do not address the combination of acting and planning. 

Recent approaches to robotic planning combine discrete task planning and continuous motion planning frameworks \cite{lagriffoul2012constraint,erdem2011combining,srivastava2014combined} pre-sampling grasps and placements producing a family of possible high level plans. These approaches use hierarchical architectures, but do not consider the continual update of the current plan. 

Other approaches\cite{levihn2013foresight} consider two types of replanning: \emph{aggressive replanning}, where replanning is done after every action; and \emph{selective replanning}: where  replanning is done whenever a  change in the environment happens that enables a new path to the goal, that is shorter than the existing plan by a given threshold. In our approach we  replan when needed. By using  continually hierarchical monitoring, we are able to monitor the part of the environment that is relevant for  goal satisfaction, disregarding  environmental changes that do not affect our plan. This enables us to plan and act in highly dynamic environments.  


The Hybrid Backward Forward (HBF) algorithm~\cite{garrettbackward} was proposed as an action planner in infinite state space. HBF is a forward search in state space, starting at the initial state of a complete domain, repeatedly selecting a state that has been visited and an action that is applicable in that state, and computing the resulting state, until a state satisfying a set of goal constraints is reached. One advantage of this approach lies in the restriction to the set of useful actions, building a so-called \emph{reachability graph}. A backward search algorithm builds the reachability graph working backward from the goal's constraints, using them to drive sampling of actions that could result in states that satisfy them. Thus, HBF enables us to deal with infinite state space, but the resulting plan is  static, and  does not address issues related to  acting and planning in dynamic environments.


When it comes to using planning to create BTs there is almost no previous work.
The closest approach is the ABL language~\cite{mateas2002behavior}.
ABL planning was created  for use in the dialogue game Fa\c{c}ade, to automatically generate complex structures from a repository of simpler ones. The structures of ABL were predecessors of BTs, including the return statuses \emph{success} and \emph{failure}, but not \emph{running}. This made reactivity much less straightforward and explicit constructions of so-called wait actions were used to respond to actions.
Furthermore, ABL planning depended on a repository of hand-made structures, whereas our approach automatically creates BTs from  a set of simple actions and their pre- and postconditions. 
Thus, ABL planning is not reactive and does not address the problems investigated in this paper.


\section{Background: Behavior Trees}
\label{sec:background}

In this section we briefly describe BTs, and refer the interested reader to the  more detailed description that can be found in~\cite{BTBook}.
A BT can be seen as a graphical modeling language and a representation for execution of actions based on conditions and observations in a system. 

\begin{figure}[h]
\centering
\includegraphics[width=0.6\columnwidth]{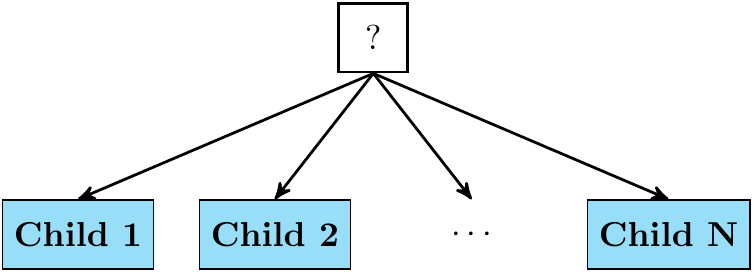}
\caption{Graphical representation of a fallback node with $N$ children.}
\label{bg.fig.sel}
\end{figure}

\begin{figure}[h]
\centering
\includegraphics[width=0.6\columnwidth]{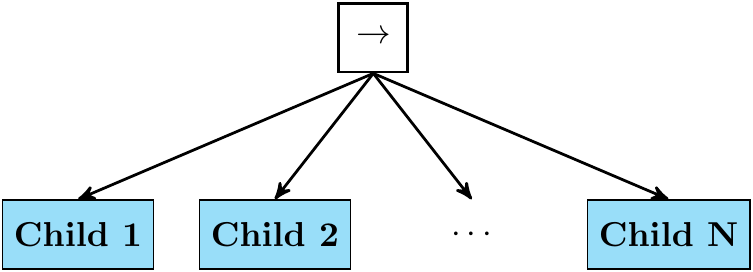}
\caption{Graphical representation of a sequence node with $N$ children.}
\label{bg.fig.seq}
\end{figure}

Formally, a BT is a directed rooted tree where each node is either a control flow node or an execution node, see below. 
We use the standard definitions of  \emph{parent} (the node above) and \emph{child} (the nodes below). The root is the single node without parents, whereas all other nodes have one parent. The control flow nodes have one or more children, and the execution nodes have no children. Graphically, the children of nodes are placed below it. The children nodes are executed in the order from left to right, as shown in Figures~\ref{bg.fig.sel}-\ref{bg.fig.seq}.

The execution of a BT begins from the root node. It sends \emph{ticks}~\footnote{A tick is a signal that allows the execution of a child} with a given frequency to its child. When a parent sends a tick to a child, the child can be executed. The child returns to the parent a status \emph{running} if its execution has not finished yet, \emph{success} if it has achieved its goal, or \emph{failure} otherwise.\\ 
There are four types of control flow nodes (fallback, sequence, parallel, and decorator) and two execution nodes (action and condition). Below we describe the execution of the nodes used in this paper.

\paragraph*{Fallback}
The fallback\footnote{Fallbacks are sometimes also called Selectors} node ticks its children from the  left, returning success (running) as soon as it finds a child that returns success (running). It returns failure only if all the children return failure. When a child returns running or success, the fallback node does not tick the next child (if any).
The fallback node is graphically represented by a box with a ``?", as in Figure~\ref{bg.fig.sel}.

\paragraph*{Sequence}
The sequence node ticks its children from the  left, returning failure (running) as soon as it finds a child that returns failure (running). It returns success only if all the children return success. When a child return running or failure, the sequence node does not tick the next child (if any). The sequence node is graphically represented by a box with a ``$\rightarrow$", as in Figure~\ref{bg.fig.seq}.

\begin{figure}[h]
       ~ 
        \begin{subfigure}[b]{0.3\columnwidth}
                \centering
                \includegraphics[width=0.4\columnwidth]{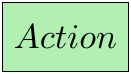}
                \caption{}
                \label{bg.fig.act}              
        \end{subfigure}
        \hspace{2cm}
        \begin{subfigure}[b]{0.3\columnwidth}
                \centering
                \includegraphics[width=0.7\columnwidth]{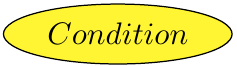}
                \caption{}
                \label{bg.fig.cond}
        \end{subfigure}
        \caption{Graphical representation of (a) an action and (b)  a condition node.}
\end{figure}
\paragraph*{Action}
The action node performs an action, returning success if the action is completed and failure if the action cannot be completed. Otherwise it returns
 running.
  An action node is shown in Figure~\ref{bg.fig.act} 
\paragraph*{Condition}
The condition node checks if a condition is satisfied or not, returning success or failure accordingly.  The condition node never returns running. A condition node is shown in Figure~\ref{bg.fig.cond} 

To get familiar with the BT notation, and prepare for the coming sections, we look at a BT plan addressing the simple example 
in Section~\ref{sec:introduction}. The BT was created using the proposed approach as will be explained in Section~\ref{sec:proposed_approach}, but for now we just focus on how it is executed.



\begin{example}
\label{ex:simple}
\label{PA.exa.simple}

The robot in Figure~\ref{IN.fig.front} is given the task to move the green cube into the rectangle marked GOAL. Ignoring the presence of the red sphere, a reactive plan BT can be found in Figure~\ref{PA.fig.it4}. Each time step, the root of the BT is ticked. The root is a fallback which ticks is first child, the condition $o_c \in GoalRect$ (cube on goal). If the cube is indeed in the rectangle we are done, and the BT returns Success.

If not, the second child, a sequence, is ticked. The sequence ticks its first child, which is a fallback, which again ticks its first child, the condition 
$h = c$ (object in hand is cube). If the cube is indeed in the hand, the condition returns success, its parent, the fallback returns success, and its parent, the sequence ticks its second child, which is a different fallback, ticking its first child which is the condition $o_r \in \mathcal{N}_{p_g}$ (robot in the neighborhood of $p_g$). If the robot is in the neighborhood of the goal, the condition and its fallback parent returns success, followed by the sequence ticking its third child, the action $Place(c,p_g)$ (place cube in a position $p_g$ on the goal), and we are done.

If $o_r \in \mathcal{N}_{p_g}$ does not hold, the action $MoveTo(p_g,\tau_g)$ (move to position $p_g$ using the trajectory $\tau_g$) is executed, given that the trajectory is free $\tau \subset C_{ollFree}$.
Similarly, if the cube is not in the hand, the robot does a $MoveTo$ followed by a $Pick(c)$ after checking that the 
hand is empty, the robot is not in the neighborhood of $c$ and that the
corresponding trajectory is free.

We conclude the example by noting that the BT is ticked every timestep, e.g. every 0.1 second. Thus, when actions return running (i.e. they are not finished yet) the return status of running is progressed up the BT and the corresponding action is allowed to control the robot. However, if e.g., the cube slips out of the gripper, the condition $h = c$ instantly returns failure, and the robot starts checking if it is in the neighborhood of the cube or if it has to move before picking it up again.
\end{example}

\section{Problem Formulation}
\label{sec:problem}
Below we will describe the main problem addressed in this paper.

\begin{table}
\normalsize
 \centering
\begin{tabular}{ c | c | c }
   Actions & Preconditions & Postconditions \\
  \hline			
 \rule{0pt}{3ex}   $A _1$& $C^{Pre}_{11} $,  $C^{Pre}_{12} $, \ldots & $C^{Post}_{11} $,  $C^{Post}_{12}$, \ldots  \\
  \rule{0pt}{3ex}  $A _2$& $C^{Pre}_{21} $,  $C^{Pre}_{22} $, \ldots & $C^{Post}_{21} $,  $C^{Post}_{22}$, \ldots \\
  $\vdots$ &$\vdots$ & $\vdots$ \\
  \hline  
\end{tabular}
\caption{The input to Problem 1 is a set of actions and corresponding pre- and post conditions, as illustrated above.}
\label{table:actions}
\end{table}

\begin{problem}
\label{paragraph:problem}
Given a set of actions, with corresponding preconditions and postconditions, as in Table~\ref{table:actions}, as well as a set of goal conditions, $C^{Goal}_{1}, C^{Goal}_{2},...$  create a BT
that strives to satisfy the goal conditions.
The BT should be reactive to changes brought about by external agents in the following  senses: 

First, if an external agent reverses an action executed by the main agent, such as taking an item from the agent and putting it on the floor, the main agent should pick it up again without having to replan in terms of expanding the BT.

Second, if an external agent carries out an action that the main agent was planning to do, such as opening a door, the main agent should take advantage of this fact, and traverse the door without trying to open it, and without having to replan in terms of expanding the BT.

Third, if there are several actions that result in a common post condition, the BT should include these so that if one fails due to external conditions, the other ones can be tried instead, without having to replan in terms of expanding the BT.

Finally, the BT should be able to be expanded during runtime. If, e.g., a condition that was earlier assumed to hold turns out to not hold, actions for achieving this conditions can be added on the fly.
\end{problem}

\section{Proposed Approach}
\label{sec:proposed_approach}

Formally, the proposed approach is described in Algorithms \ref{PA.alg.main}, \ref{PA.alg.update}   and    \ref{PA.alg.getcond},
but before going into detail, we will
first  describe the two main ideas of the algorithms, i.e., creating atomic BTs and iteratively replacing failed conditions with these atomic BTs.
Then we will see how the algorithms are applied to the problem described in \emph{Example 1},
to iteratively create the BTs of Figure \ref{PA.fig.it1to4}.
Finally, we discuss the key steps in more detail.

\subsection{Atomic BTs for each postcondition}
\label{backchaining}

The first step of the algorithm converts the list of actions in Table~\ref{table:actions} into a list of atomic BTs, each aimed at satisfying a given condition, but invoking the actions only when the condition is not met. This construction is what enables the reactivity needed in Problem~\ref{paragraph:problem}.

\begin{figure}[h!]
\centering
\includegraphics[width=\columnwidth]{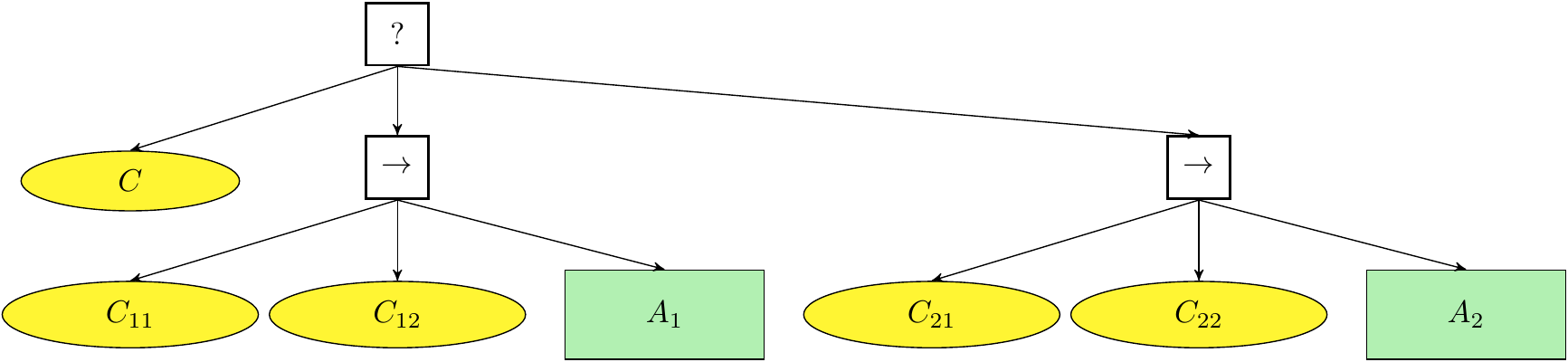}
\caption{General format of an atomic BT. The Postcondition $C$ can be achieved by either one of actions $A_1$ or $A_2$, which have Preconditions $C_{11},C_{12}$ and $C_{21},C_{22}$ respectively.)}
\label{design:fig:back_chaining_general}
\end{figure}

Assume that the table includes a postcondition $C$ that can be achieved by either action $A_1$ or action $A_2$, that in turn have preconditions $C_{11},C_{12}$ and $C_{21},C_{22}$ respectively. Then we create an atomic BT
aimed at achieving the condition $C$ by composing the actions and conditions in the generic way displayed in Figure~\ref{design:fig:back_chaining_general}, i.e., each action $A_i$ in sequence after its preconditions $C_{ij}$, and these sequences in a fallback composition after the main condition $C$ itself.
Finally we create similar BTs for each postcondition $C$ of  Table~\ref{table:actions}.

\begin{remark}
Note that the order of both actions $A_1, A_2$ and preconditions $C_{11},C_{12}$ were  arbitrary in  the BT of Figure~\ref{design:fig:back_chaining_general}. We will later enable the re-ordering of the pre-conditions based on so-called conflicts. 
For example, if you want to transport object X and move object Y out of the way, you have to put down object X before using the arm to move object Y. Thus it makes sense to re-order the conditions so that moving object Y is done before transporting X.
\end{remark}

\begin{remark}
One can also consider reordering the actions, based  on success probabilities and executions times,
as it often makes sense to try quick solutions before slow ones, and dependable ones before risky ones.
\end{remark}

\subsection{Iteratively expanding a BT from the Goal Conditions}
\label{backchainingII}
Having a list of atomic BTs we can now iteratively build
a deliberative BT by starting with a minimalistic BT, made up of a single sequence composition of all the goal conditions.
Then we execute this BT. If it returns success all conditions are met and we are done.
If not we replace each condition that returned failure with the corresponding atomic BT, of the form shown in Figure~\ref{design:fig:back_chaining_general}.
Note that this BT returns success immediately if the condition is met, but tries to satisfy the condition if it is not.
The new BT is again executed. As long as it returns running we let it run. If it succeeds we are done, and if it fails
we replace the failing condition with a new atomic BT. In the next paragraph we will see that this is how the BTs of Figure~\ref{PA.fig.it1to4} where created.

\begin{algorithm2e}[h!]
\caption{Main Loop, finding conditions to expand and resolve conflicts} 
 \label{PA.alg.main}
\DontPrintSemicolon
%
\SetKwProg{myalg}{algorithm2e}{}{}
 $\mathcal{T} \gets \emptyset$ \\ 
\For{$c$ in $\mathcal{C}_{goal}$} {
 $\mathcal{T} \gets $SequenceNode($\mathcal{T}$, $c$)}
\While{True\label{PA.alg.main.while}}{
$T\gets$\FuncSty{RefineActions($\bt$)} \label{PA.alg.main.refine}\\
  \Do{\label{PA.alg.main.do} $r \neq \mbox{\emph{Failure}}$ \label{PA.alg.main.fail}} {
    $r \gets$ \FuncSty{Tick($T$)} \label{PA.alg.main.fail2}}
 $c_f \gets$ \FuncSty{GetConditionToExpand($\bt$)}  \label{PA.alg.expand.getcon}\\
 $\mathcal{T}, \mathcal{T}_{new\_subtree}\gets$ \FuncSty{ExpandTree($\bt$,$c_f$)}\label{PA.alg.main.expand} \\
   \While{ $ \mbox{Conflict}(\bt)$  \label{PA.alg.main.feas}} {  $\mathcal{T}\gets$ \FuncSty{IncreasePriority($\mathcal{T}_{new\_subtree}$)} \label{PA.alg.main.incprio} } }
 
\end{algorithm2e}

\begin{algorithm2e}[h]
\caption{Replace failed condition with new Atomic BT }
      \label{PA.alg.update}
  \SetKwFunction{algo}{ExpandTree}
  \SetKwFunction{proc}{proc}
  \SetKwProg{myalg}{Function}{}{}
  \myalg{\algo{$\mathcal{T}$, $c_f $}}{


 $A_T \gets $ \FuncSty {GetAllActTemplatesFor($c_f$) \label{PA.alg.expand.getact}}  \\
 	$\mathcal{T}_{fall} \gets c_f$\\ 
\For{ $a$ in $A_T$} {
	$\mathcal{T}_{seq} \gets \emptyset$\\
\For{ $c_a$ in $a.con$}{
 $\mathcal{T}_{seq} \gets $ SequenceNode($\mathcal{T}_{seq}$,$c_a$)   \\  
}
 $\mathcal{T}_{seq} \gets $ SequenceNode($\mathcal{T}_{seq}$,$a$) \\ 
 $\mathcal{T}_{fall} \gets $ FallbackNode($\mathcal{T}_{fall}$,$\mathcal{T}_{seq}$) \\ 
}

 $\mathcal{T} \gets$  Substitute($\mathcal{T}$,$c_f$,$\mathcal{T}_{fall}$)\\
 \Return{$\mathcal{T}$, $\mathcal{T}_{fall}$}
 }
 
 \end{algorithm2e}

\begin{algorithm2e}[h]

\caption{Get Condition to Expand}
      \label{PA.alg.getcond}
  \SetKwFunction{algo}{GetConditionToExpand}
  \SetKwFunction{proc}{proc}
  \SetKwProg{myalg}{Function}{}{}
  \myalg{\algo{$\mathcal{T}$}}{
	\For{$c_{next} $ in $\FuncSty{GetConditionsBFS()}$}
	{
		\If{$c_{next}.status =\mbox{\emph{Failure}}$ \textbf{and} $c_{next} \notin \mbox{ExpandedNodes}$ \label{PA.alg.getcond.notexpanded}}
		{
		\mbox{ExpandedNodes}.\FuncSty{push\_back($c_{next}$)}
			\Return{$c_{next}$}		
		}
	} 
	\Return{$None$}\label{PA.alg.getcond.nocond}
 }

\end{algorithm2e}

\subsection{Algorithm Example Execution}
\begin{figure}[t]
        \centering
          \begin{subfigure}[b]{0.25\columnwidth}
                \centering
\includegraphics[width=1\columnwidth]{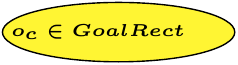}
                \caption{The initial BT.}
                \label{PA.fig.it0}
        \end{subfigure}%
        
        \begin{subfigure}[b]{0.5\columnwidth}
                \centering
\includegraphics[width=1\columnwidth]{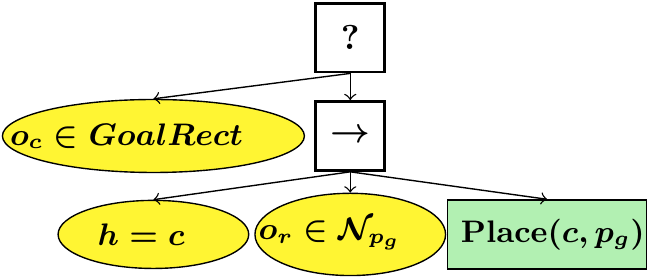}
                \caption{BT after one iteration.}
                \label{PA.fig.it1}
        \end{subfigure}%
       ~ 
        \begin{subfigure}[b]{0.5\columnwidth}
                \centering
\includegraphics[width=1\columnwidth]{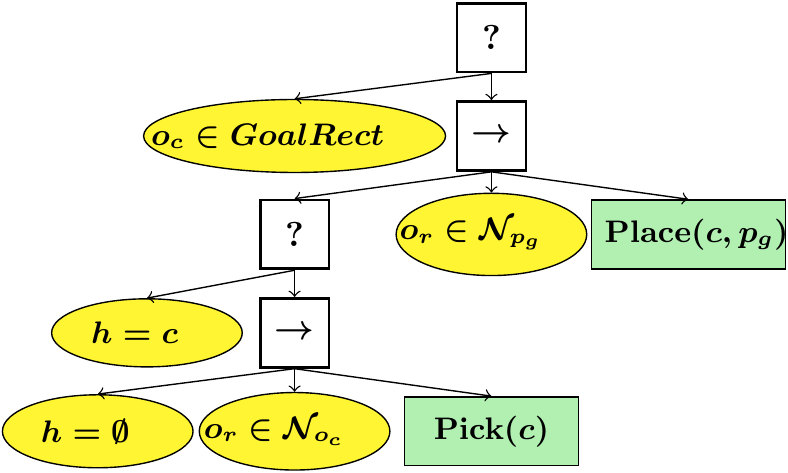}
                \caption{BT after two iterations.}
                \label{PA.fig.it2}              
        \end{subfigure}
        ~ 

        \centering
        \begin{subfigure}[b]{0.45\columnwidth}
                \centering
\includegraphics[width=1\columnwidth]{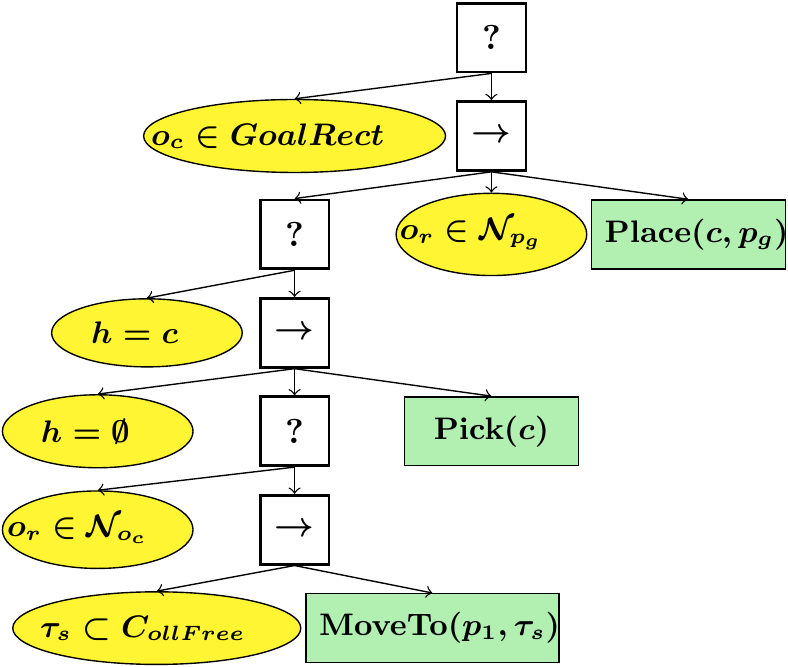}
                \caption{BT after three iterations.}
                \label{PA.fig.it3}
        \end{subfigure}%
       ~ 
        \begin{subfigure}[b]{0.55\columnwidth}
                \centering
\includegraphics[width=1\columnwidth]{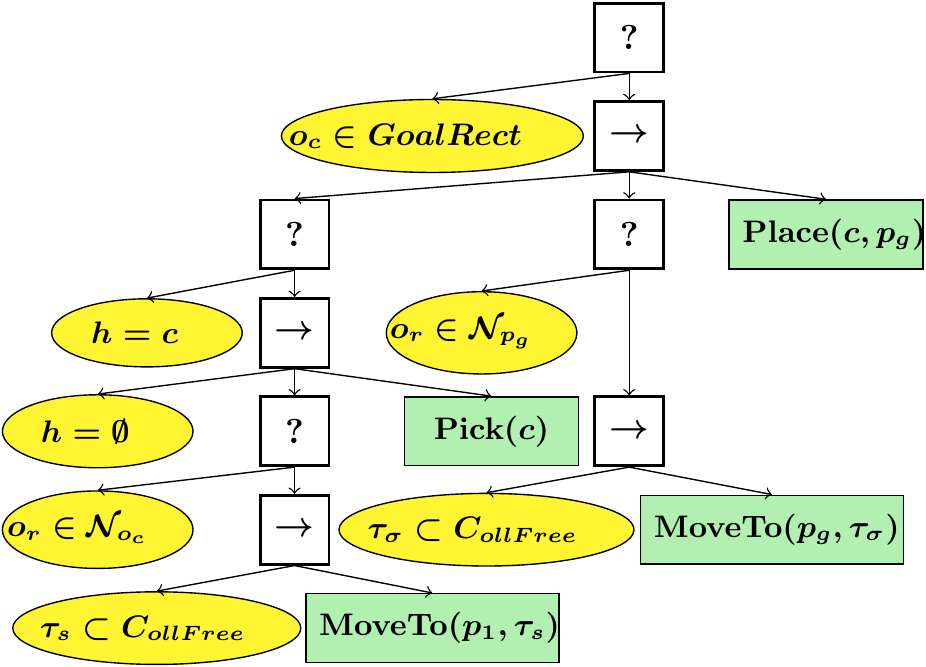}
                \caption{BT after four iterations. Final Tree}
                \label{PA.fig.it4}              
        \end{subfigure}
        
        \caption{BT updates during the execution.}
         \label{PA.fig.it1to4}              
        ~ 
\end{figure}

This example starts with a single goal conditions, shown in Figure~\ref{PA.fig.it0}.
Running Algorithm~\ref{PA.alg.main} we have the set of goal constraint  $\mathcal{C}_{goal} = \{o_c \in \{\mbox{GoalRect}\} \}$, thus the initial BT
is composed of a single condition
 $\bt = (o_c \in \{\mbox{GoalRect}\})$, as shown in Figure~\ref{PA.fig.it0}.
 The first iteration of the loop starting on Line \ref{PA.alg.main.while} of Algorithm~\ref{PA.alg.main} now produces the next BT shown in Figure~\ref{PA.fig.it1}, and the second iteration produces the BT in Figure \ref{PA.fig.it2} and so on until the final BT in Figure \ref{PA.fig.it4}.

In detail, running $\bt$ on Line \ref{PA.alg.main.fail2} returns a failure, since the cube is not in the goal area. Trivially, the failed condition is  $c_f=(o_c \in \{\mbox{GoalRect}\})$,
and a call to ExpandTree (Algorithm~\ref{PA.alg.update}) is made on Line \ref{PA.alg.main.expand}. 
On Line 2 of Algorithm~\ref{PA.alg.update} we get $A_T=Place$.
Then on Line 7 and 8 a sequence $\bt_{seq}$ is created of the conditions of $Place$ (the hand holding the cube $h = c$ and the robot being near the goal area $o_r \in \mathcal{N}_{p_g}$) and $Place$ itself.
On Line 9 a fallback $\bt_{seq}$ is created of $c_f$ and the sequence above. 
Finally, a BT is returned where this new sub-BT is replacing $c_f$.
The resulting BT is shown in Figure~\ref{PA.fig.it1}. 

Note that Algorithm~\ref{PA.alg.update} describes \emph{the core principle} of the proposed approach. The BT is iteratively extended as described in Sections~\ref{backchaining} to \ref{backchainingII}. 

Running the next iteration of Algorithm~\ref{PA.alg.main}, a similar expansion of the condition $h = c$ transforms the BT in Figure~\ref{PA.fig.it1} to the BT in Fig.~\ref{PA.fig.it2}.
Then, an expansion of the condition $o_r \in \mathcal{N}_{o_c}$ transforms the BT in Figure~\ref{PA.fig.it2} to the BT in Figure~\ref{PA.fig.it3}.
Finally, an expansion of the condition $o_r \in \mathcal{N}_{p_g}$ transforms the BT in Figure~\ref{PA.fig.it3} to the BT in Figure~\ref{PA.fig.it4},
and this BT is able to solve \emph{Example 1}.

%
%

\subsection{The Algorithm Steps in Detail}

\subsubsection{Refine Actions (Algorithm~\ref{PA.alg.main} Line~\ref{PA.alg.main.refine})}
 This process implements an action refinement as described in~\cite{Nau15Challenges}, that is, we map template actions and conditions (e.g. $Place(c,P_g)$) in to grounded actions and conditions (e.g. $Place(c,[0,0])$). Grounded actions can be executed by the robot. We assume that a valid action refinement always exists, handling cases where it does not is beyond the scope of this paper.
\subsubsection{Get Deepest Failed Condition and Expand Tree (Algorithm~\ref{PA.alg.main} Lines~9 and~10) }

If the BT returns failure, Line~\ref{PA.alg.expand.getcon} finds the deepest condition returning failure. This will then be expanded, as shown in the example of Figure~\ref{PA.fig.it1to4}.
$\bt$ is expanded until it can perform an action (i.e. until $\bt$ contains an action template whose condition are supported by the initial state). 
If there exists more than one valid action that satisfies a condition, their respective trees (sequence composition of the action and its conditions) are collected in a fallback composition, which implements the different options the agent has to satisfy such a condition. 
Note that 
at this stage we do not investigate which action is the optimal one. As stressed in \cite{ghallab2016automated} the cost of minor mistakes (e.g. non optimal actions execution) is often much lower than the cost of extensive modelling, information gathering and thorough deliberation needed to achieve optimality.

\subsubsection{Conflicts and Increases in Priority (Algorithm~\ref{PA.alg.main} Lines~\ref{PA.alg.main.feas} and~\ref{PA.alg.main.incprio} )}
Similar to any STRIPS-style planner, adding a new action  in the plan can cause a \emph{conflict} (i.e. the execution of this new action creates a missmatch between effects and preconditions the progress of the plan). In our framework, this situation is checked in Algorithm~\ref{PA.alg.main}  Line~\ref{PA.alg.main.feas} by analyzing the conditions of the new action added with the effects of the actions that the subtree executes before executing the new action. If this effects/conditions pair is in conflict, the goal will not be reached.  An example of this situation is described in Example~\ref{PA.exa.complex} below.


Again, following the approach used in STRIPS-style planners, we resolve this conflict by finding the correct action order. Exploiting the structure of BTs we can do so by moving the tree composed by the new action and its condition leftward (a BT executes its children from left to right, thus moving a sub-tree leftward  implies that it will be executed earlier). If it is the leftmost one, is means that it must be executed before its parent (i.e. it must be placed at the same depth of the parent but to its left). This operation is done in Algorithm~\ref{PA.alg.main}  Line~\ref{PA.alg.main.incprio}. We incrementally increase the priority of the subtree in this way, until we find a feasible tree. 
We assume a non conflicting order exists. Cases were this is not the case can be constructed, but these problems are beyond the scope of this paper.

\begin{figure}[t]
        \centering
        \begin{subfigure}[b]{0.5\columnwidth}
                \centering
\includegraphics[width=1\columnwidth]{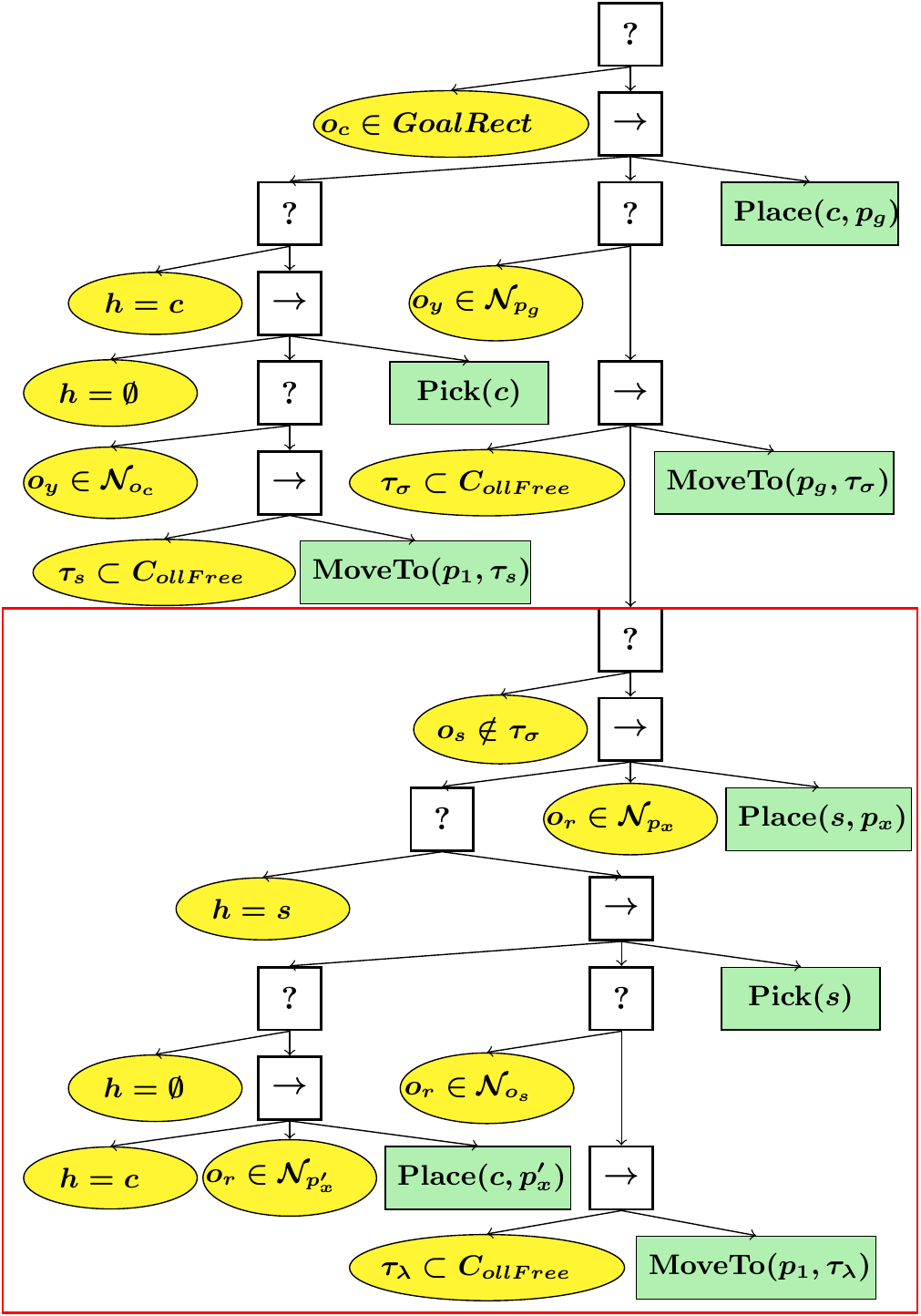}
                \caption{Unfeasible expanded tree. The new subtree is highlighted in red.}
                \label{PA.fig.it7}
        \end{subfigure}\\
       ~ 
        \begin{subfigure}[b]{1\columnwidth}
                \centering
\includegraphics[width=1\columnwidth]{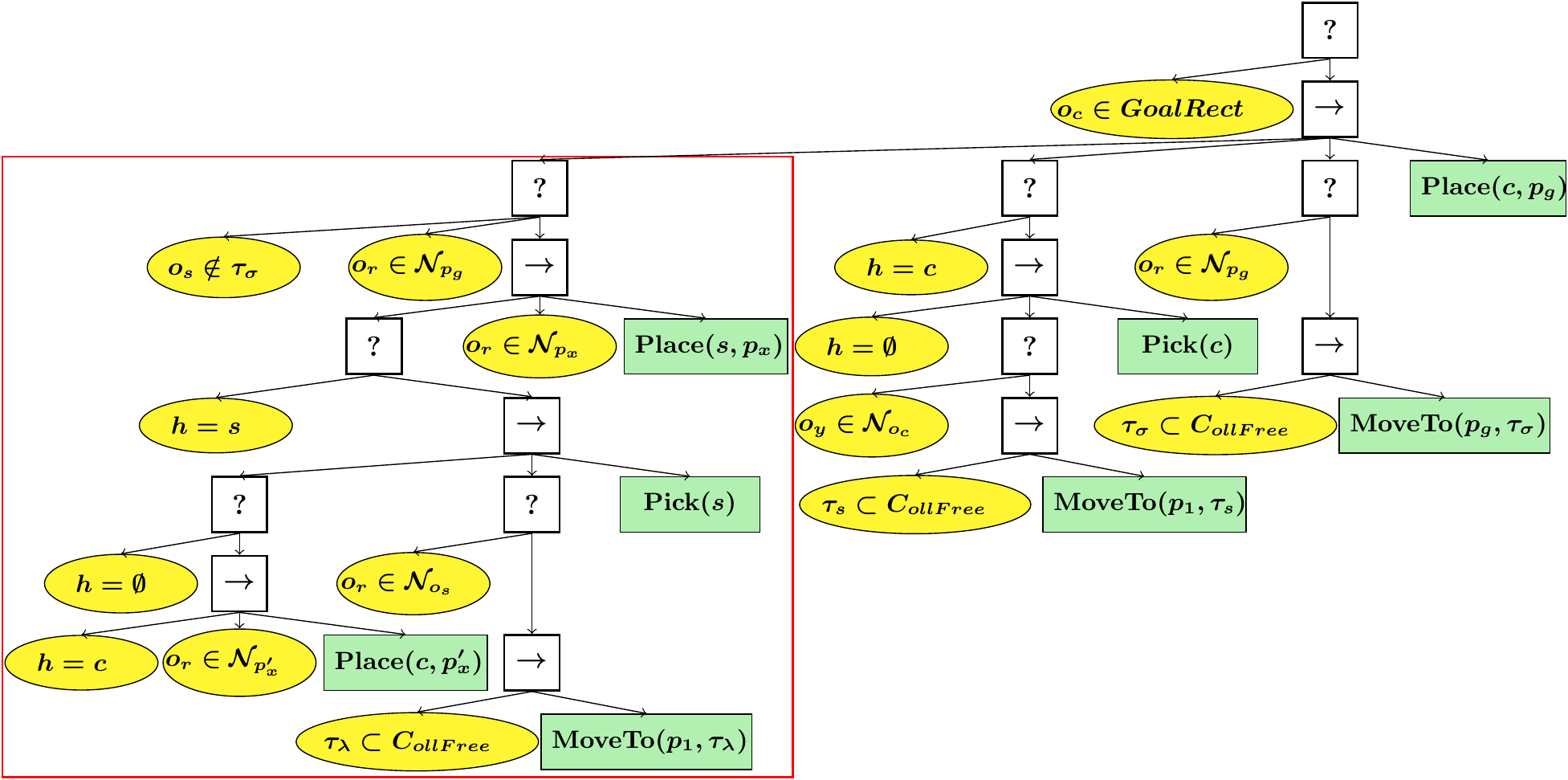}
                \caption{Expanded Feasible subtree.}
                \label{PA.fig.it8}              
        \end{subfigure}
        ~ 
        \caption{Steps to increase the priority of the new subtree added in Example~\ref{PA.exa.complex}.}
        ~ 
\end{figure}

\begin{example}
\label{PA.exa.complex}

Here we show a more complex example including a conflict, and illustrating the continual deliberative plan and act cycle. This example is an extension of Example~\ref{PA.exa.simple} where, due to the dynamic environment, the robot has to replan.


Consider the execution of the final BT, Figure~\ref{PA.fig.it4} of Example \ref{PA.exa.simple}, where the robot is carrying the desired object to the goal location. Suddenly, as in Figure~\ref{IN.fig.front} (a), an object $s$ obstructs the (only possible) path. Then the condition $\tau \subset C_{ollFree}$  returns failure and Algorithm \ref{PA.alg.main} expands the tree accordingly (Line \ref{PA.alg.main.expand}) as in Figure \ref{PA.fig.it7}.

The new subtree has as condition $h = \emptyset$ (no objects in hand) but the effect of the left branch (i.e. the main part in Figure \ref{PA.fig.it4}) of the BT is $h = c$ (cube in hand) (i.e. the new subtree will be executed if and only if $h = c$ holds). Clearly  the expanded tree has a conflict (Algorithm~\ref{PA.alg.main} Line \ref{PA.alg.main.feas}) and the priority of the new subtree is increased  (Line~\ref{PA.alg.main.incprio}), until the expanded tree is in form of Figure~\ref{PA.fig.it8}. Now the BT is free from conflicts as the first subtree has as effect $h = \emptyset$  and the second subtree has a condition $h = \emptyset$. Executing the tree the robot approaches the obstructing object, now the condition $h = \emptyset$ returns failure and the tree is expanded accordingly, letting the robot drop the current object grasped, satisfying $h = \emptyset$, then it picks up the obstructing object and places it outside the path. Now the condition $\tau \subset C_{ollFree}$ finally returns success. The robot can then again approach the desired object and move to the goal region and place the object in it.  
A video showing the entire execution is publicly available~\footnote{https://youtu.be/wTQNVW38u4U}.

\end{example}

\subsubsection{Get All Action Templates}

Let's look again at \emph{Example~1} and see how the BT in Figure~\ref{PA.fig.it4} was created using the proposed approach.

In this example, the action templates are summarized below with pre- and post-condition:
\begin{equation*}
\begin{aligned}
&\mbox{MoveTo}(p,\tau)   \\   
&\mbox{pre\;}:\tau \subset C_{ollFree} \hspace{2em}  \\ 
& \\ 
&\mbox{post}:o_r = p
\end{aligned}
\begin{aligned}
&\mbox{Pick}(i)   \\   
&\mbox{pre\;}: o_r \in \mathcal{N}_{o_i} \hspace{2em} \\ 
	 &\hspace{2.3em}h  =  \emptyset \\	  
&\mbox{post}: h = i \hspace{1em} 
\end{aligned}
\begin{aligned}
&\mbox{Place}(i,p)   \\   
&\mbox{pre\;}: o_r \in \mathcal{N}_{p} \hspace{2em}  \\ 
	 &\hspace{2.3em}h = i  \\	  
&\mbox{post}: o_i = p \hspace{1em} 
\end{aligned}
\end{equation*}
where $\tau$ is a trajectory, $C_{ollFree}$ is the set of all collision free trajectories,  $o_r$ is the robot pose, $p$ is a pose in the state space, $h$ is the object currently in the end effector, $i$ is the label of the $i$-th object in the scene, and  $\mathcal{N}_{x}$ is the set of all the poses near the pose $x$. 


The descriptive model of the action \emph{MoveTo}  is parametrized over the destination $p$ and the trajectory $\tau$. It requires that the trajectory is collision free ($\tau \subset C_{ollFree}$). As effect the action \emph{MoveTo} places the robot at $p$ (i.e. $o_r = p$); The descriptive model of the action \emph{Pick} is parametrized over object $i$. It requires having the end effector free (i.e. $h  =  \emptyset$) and the robot to be in a neighbourhood $\mathcal{N}_{o_i}$ of the object $i$. (i.e. $o_r \in \mathcal{N}_{o_i}$). As effect the action Pick sets the object in the end effector to $i$ (i.e $h = i$); 
Finally, the  descriptive model of the action \emph{Place} is parametrized over object $i$ and final position $p$. It requires the robot to hold $i$, (i.e. $h = i$), and the robot to be in the neighbourhood of the final position $p$. As effect the action \emph{Place} places the object $i$ at $p$ (i.e. $o_i = p$).

\subsection{Do Algorithms 1-3 solve Problem 1?}
A solution to Problem 1 needs to be reactive in three ways.
Looking at the solution to Example 1, shown in Figure~\ref{PA.fig.it4},
we see that if the cube c is removed from the agent it will pick it up again without replanning.
We also see that if the cube is somehow placed in the hand of the agent, the agent will skip moving to the proper place and picking it up.
The example does not include postconditions that can be achieved by several actions, but it is clear from the construction of Algorithm~2 that such functionality is included.
Finally, it is clear from Example 2, and the solution in  Figure~\ref{PA.fig.it8}, that the algorithm can respond to unexpected events 
by extending the BT when necessary.           

\begin{remark}
One might want to avoid  retrying a previously failed action until a given time has passed, or circumstances have changed significantly. In that case, one adds an extra precondition before the action capturing these requirements.
\end{remark}

\section{Simulations}
\label{sec:simulations}

In this section we show how the proposed approach scales to complex problems using two different scenarios. First, a KUKA Youbot scenario, where we show the applicability of our approach on dynamic and unpredictable environments, highlighting the importance of continually planing and acting. Second, an  ABB Yumi industrial manipulator scenario, where we highlight the applicability of our approach to real world plans that require the execution of a long sequence of actions. The experiments were carried out using the physic simulator V-REP,
in-house implementations of
 low level controllers for actions and conditions and an open source BT library\footnote{http://wiki.ros.org/behavior\_tree}. 
 The action refinement algorithm used is a modified version of the one used in the HBF algorithm~\cite{garrettbackward}.

 Since capturing long reactive action sequences is difficult in pictures,
 a video showing both the scenarios below is publicly available\footnote{The videos will be available at publication, please see the uploaded video}.

%
%
%
%
\subsection{KUKA Youbot experiments}

\begin{figure}[t]

        \centering
        \begin{subfigure}[b]{1\columnwidth}
                \centering
                \includegraphics[width=0.9\columnwidth,trim={0 3cm 0 8cm},clip]{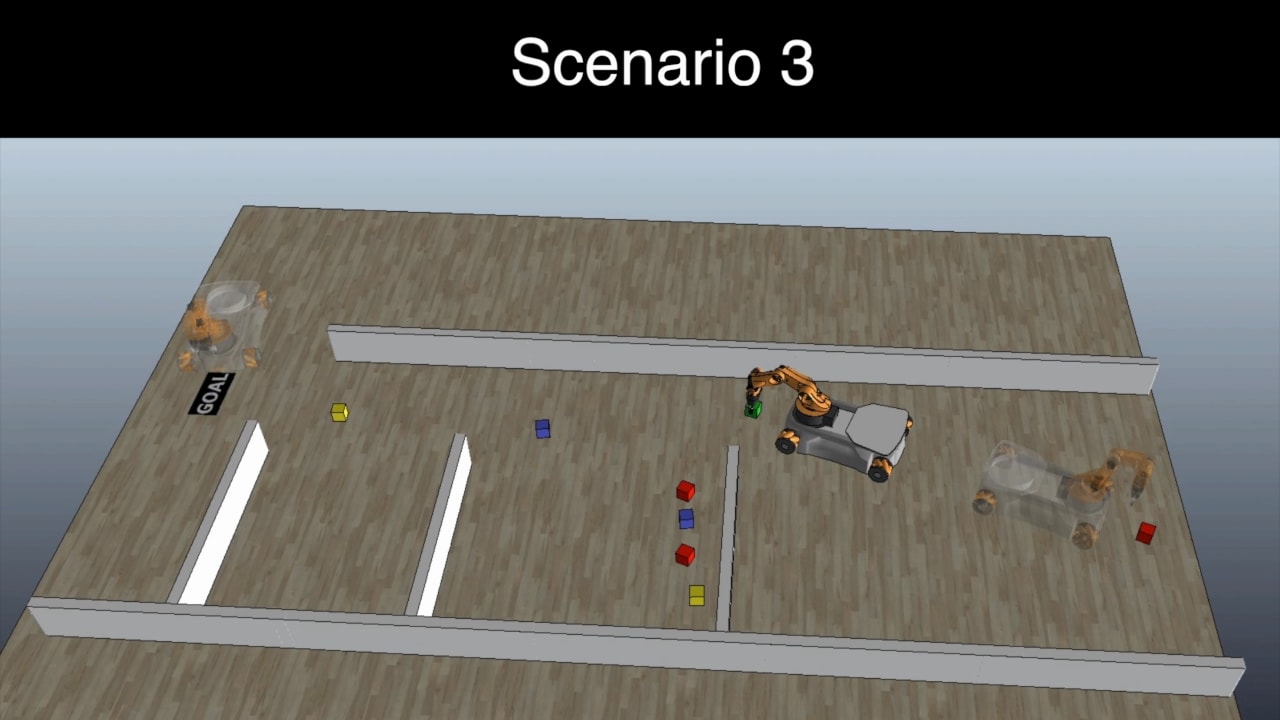}
                \caption{The robot picks up the desired object, a green cube. \\ \hspace*{1em}}
                \label{SI.fig.youbotstep1}              
        \end{subfigure}       
~

\centering
        \begin{subfigure}[b]{1\columnwidth}
                \centering
                \includegraphics[width=0.9\columnwidth,trim={0 3cm 0 8cm},clip]{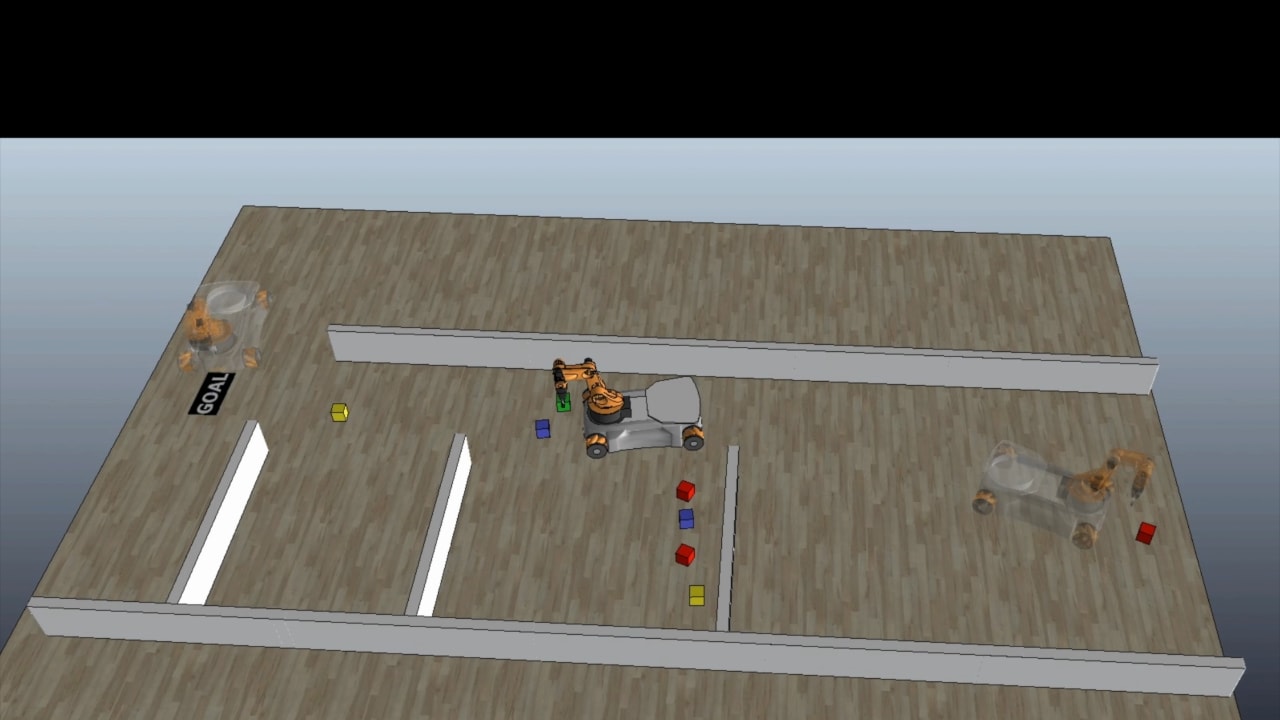}
                \caption{The blue cube obstructs the path to the goal region. The robot drops the green cube and picks up the blue cube. \\ \hspace*{1em}}
                 \label{SI.fig.youbotstep3}  
        \end{subfigure}   
        ~            
        \begin{subfigure}[b]{1\columnwidth}
                \centering
                \includegraphics[width=0.9\columnwidth,trim={0 3cm 0 8cm},clip]{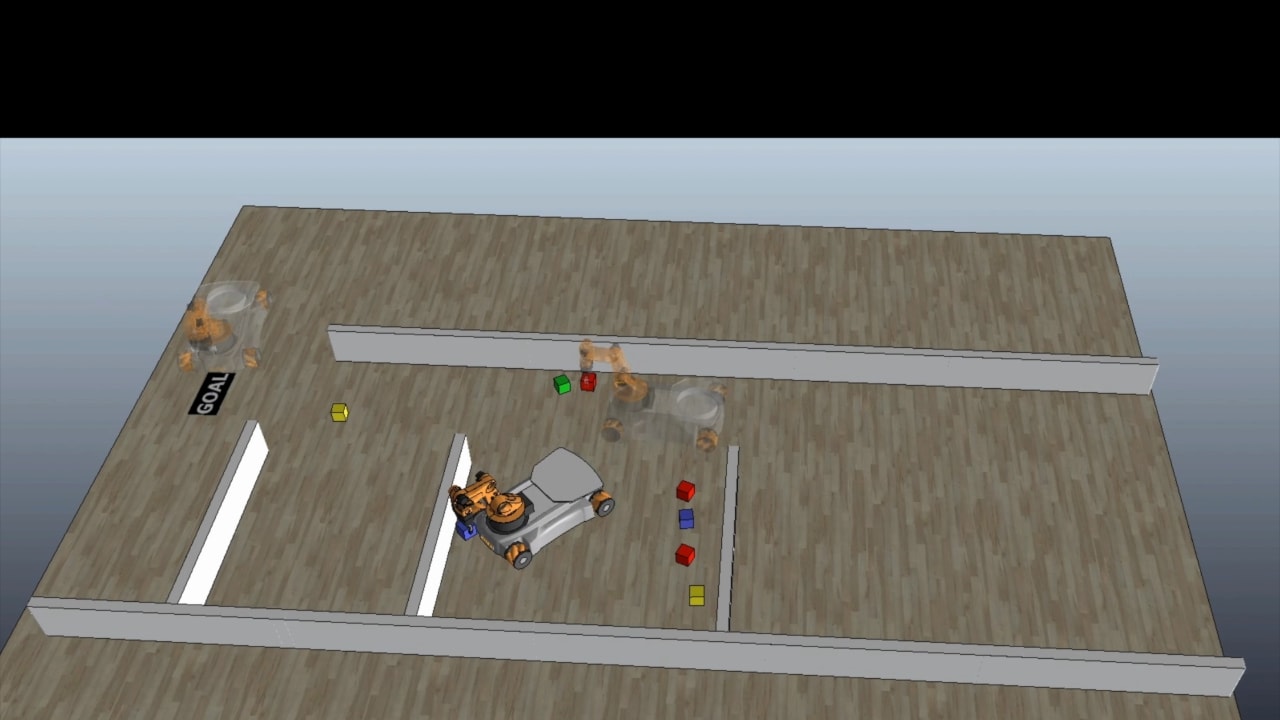}
                \caption{While the robot places the blue cube to the side of the path to the goal, an external agent places a red cube between the robot and the green cube.}
                 \label{SI.fig.youbotstep4}  
        \end{subfigure} 
~

\centering

      \begin{subfigure}[b]{1\columnwidth}
                \centering
                \includegraphics[width=0.9\columnwidth,trim={0 3cm 0 8cm},clip]{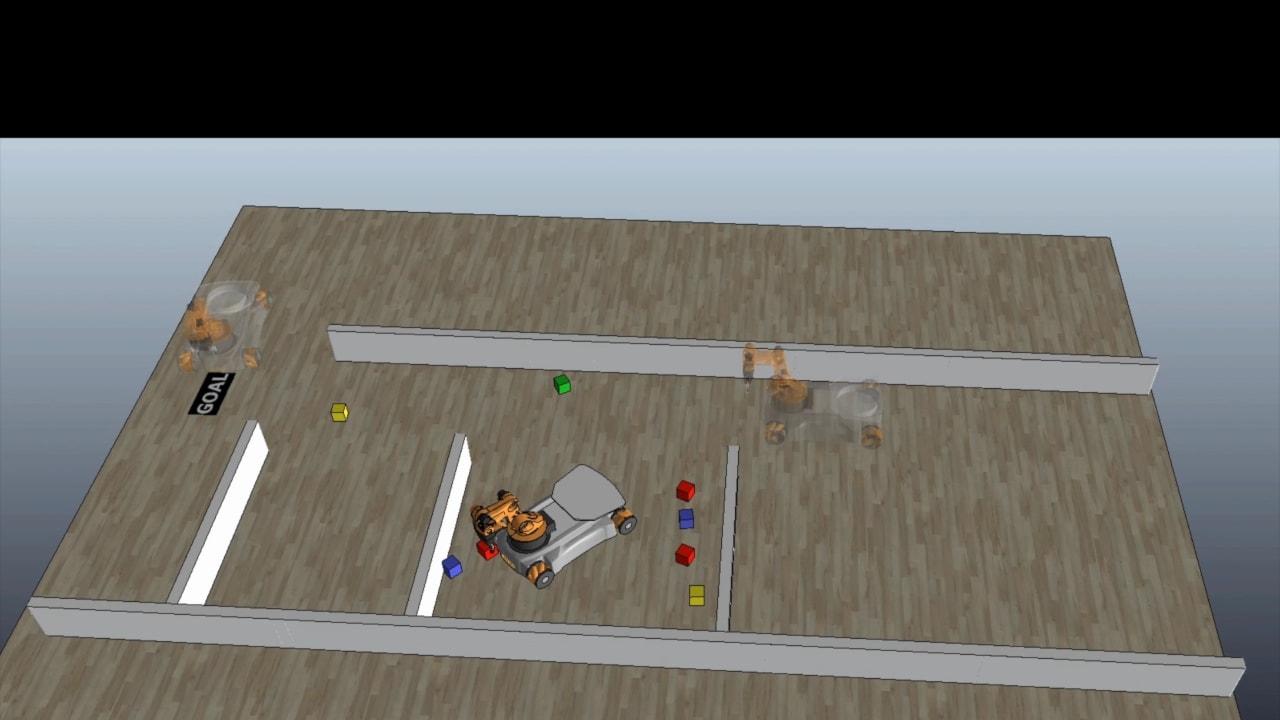}
                \caption{The robot places the red cube to the side of the path to the goal.  \\ \hspace*{1em} }
                 \label{SI.fig.youbotstep5}  
        \end{subfigure}         
        ~
              \begin{subfigure}[b]{1\columnwidth}
                \centering
                \includegraphics[width=0.9\columnwidth,trim={0 3cm 0 8cm},clip]{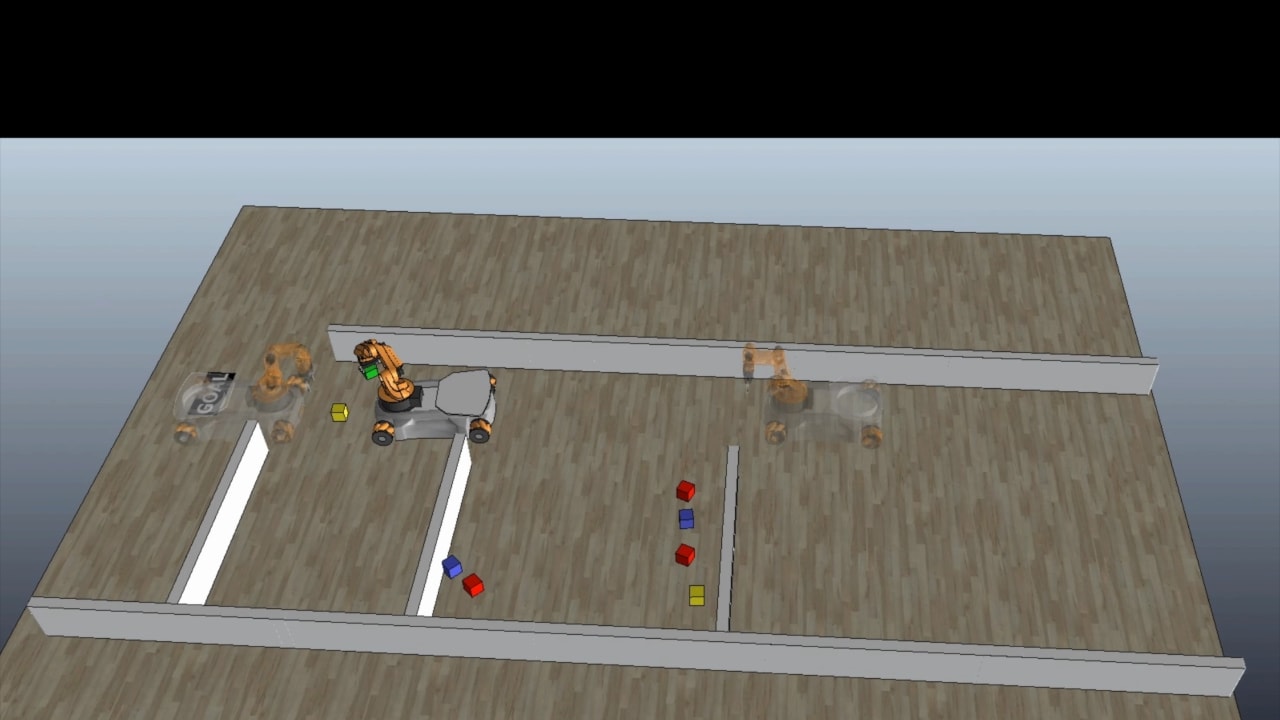}
                \caption{The yellow cube obstructs the path to the goal region. The robot drops the green cube in order to pick up the yellow cube.}
                 \label{SI.fig.youbotstep6}  
        \end{subfigure} 
        ~
        
        \centering
              \begin{subfigure}[b]{1\columnwidth}
                \centering
                \includegraphics[width=0.9\columnwidth,trim={0 3cm 0 8cm},clip]{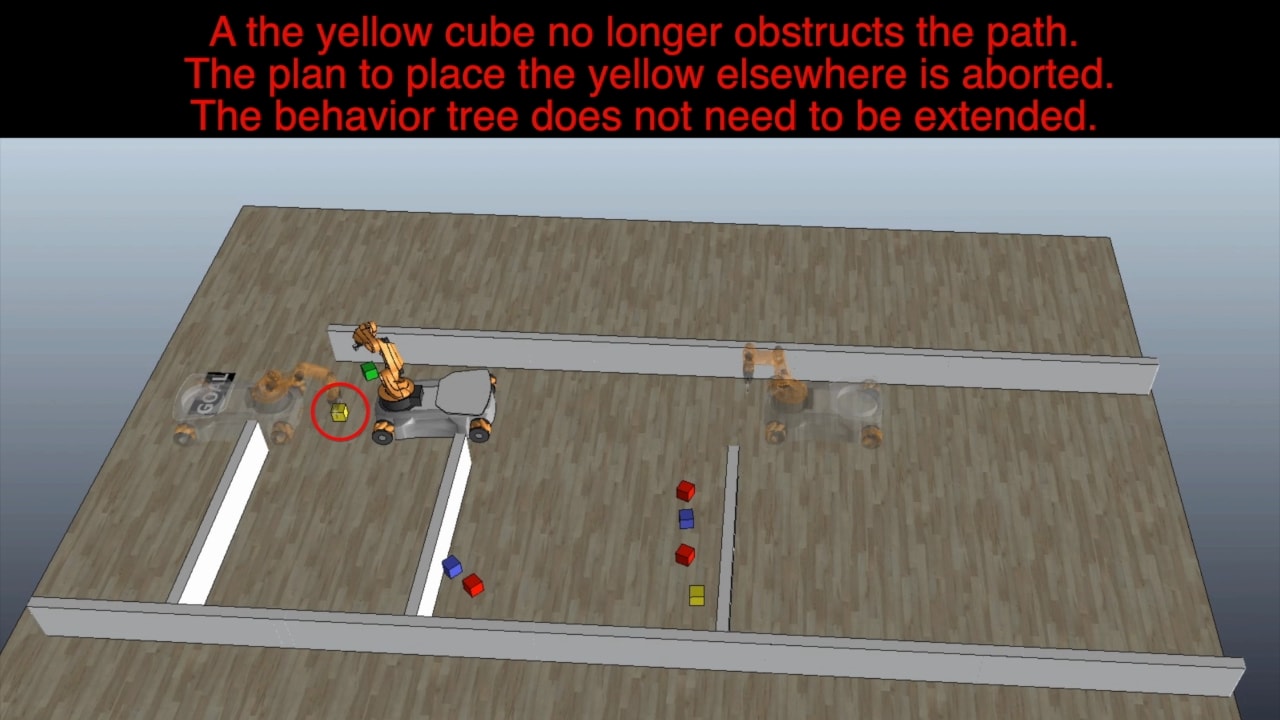}
                \caption{While the robot approaches the yellow cube, an external agent removes it.}
                 \label{SI.fig.youbotstep7}  
        \end{subfigure}      
        ~   
              \begin{subfigure}[b]{1\columnwidth}
                \centering
                \includegraphics[width=0.9\columnwidth,trim={0 3cm 0 8cm},clip]{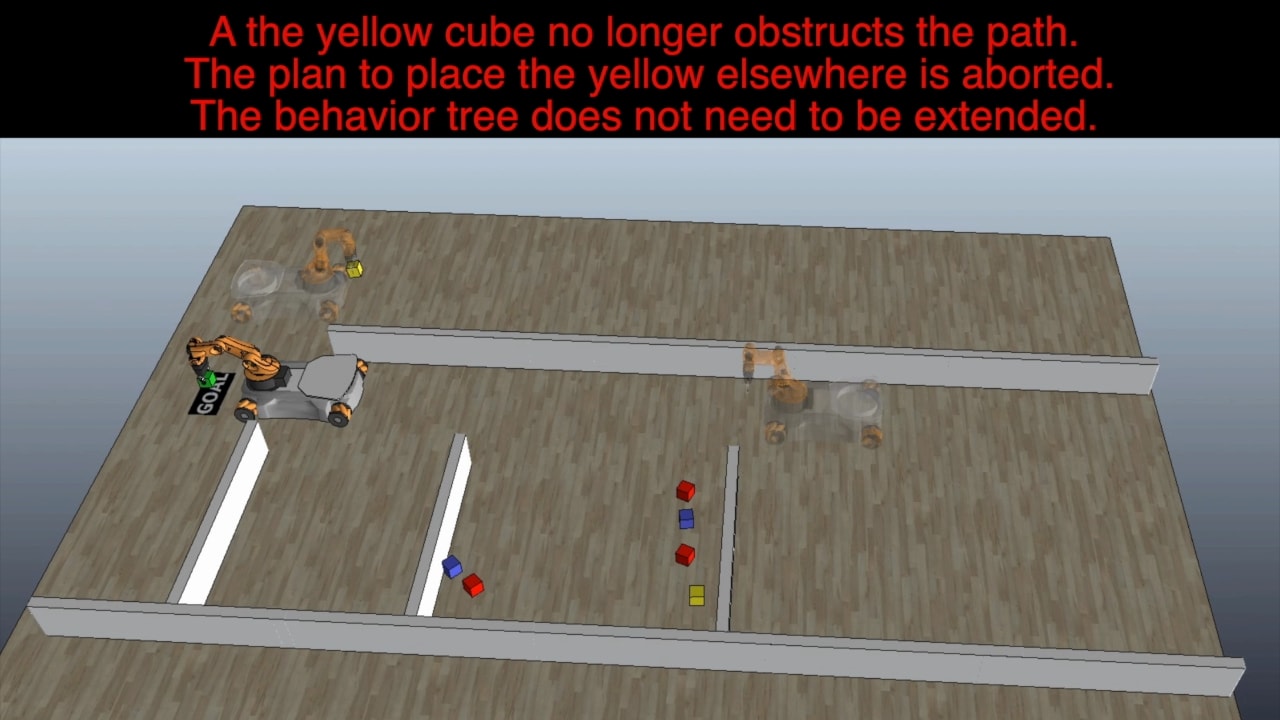}
                \caption{The robot ignores the yellow cube, picks up the green cube and places it on the goal region.\\ \hspace*{1em} }
                 \label{SI.fig.youbotstep8}  
        \end{subfigure}        
        \caption{Execution of the KUKA Youbot experiment.}
        \label{SI.fig.youscreen}
\end{figure}

In this scenario, which is an extension of Examples 1 and~2, a KUKA Youbot has to place a green cube on a goal area. The robot is equipped with a single arm with a simple parallel gripper. Additional objects may obstruct the feasible paths to the goal, and the robot has to plan when to pick and where to place to the obstructing objects. Moreover, two external agents move around in the scene and force the robot to replan by  modifying the environment.
Figure~\ref{SI.fig.youscreen} shows the planning and acting steps executed.

%
%
%
%
%
%
%
%
%
%
%
%

\subsection{ABB Yumi experiments}

\begin{figure}[h]
\centering
\includegraphics[width=\columnwidth,trim={4.5cm 3.5cm 6cm 0cm},clip]{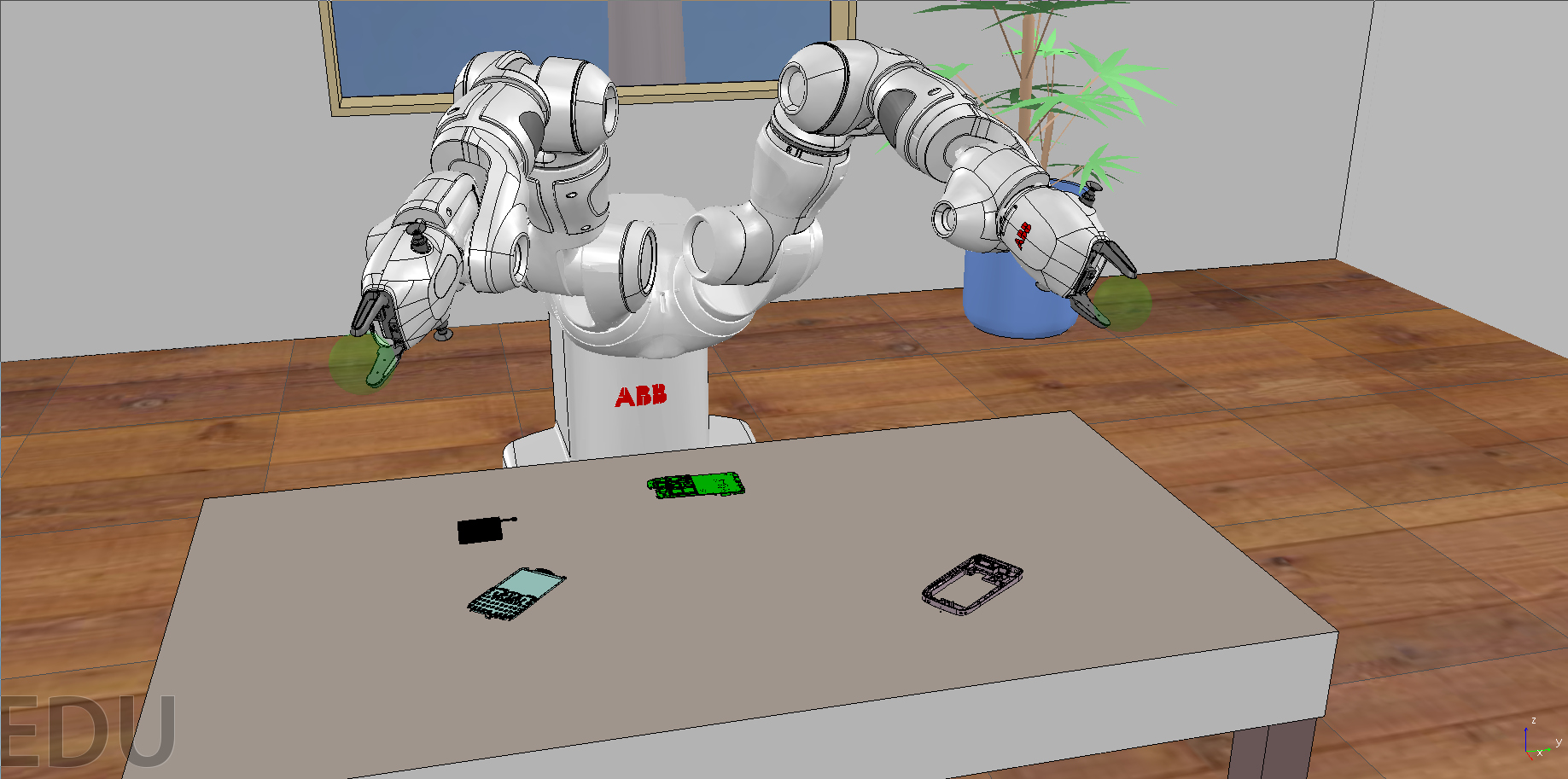}
\caption{Scenario of the ABB Yumi experiments. See video for details.}
\label{SI.fig.yum}
\end{figure}

In this scenario, an ABB Yumi has to assemble a cellphone, whose parts are scattered across a table, see Figure~\ref{SI.fig.yum}. 
The robot is equipped with two arms with simple parallel grippers, which prevents any kind of dexterous manipulation.
Some parts must be grasped in a particular position. 
For example the opening on the cellphone's chassis has to face away from the robot's arm, exposing it for the assembly.

\clearpage

However, the initial position of a part can be such that it requires multiple grasps transferring the part to the other gripper, effectively changing its orientation w.r.t the grasping gripper.

\section{Conclusions}
\label{sec:conclusions}

In this paper we proposed an approach to automatically create and update a BT using a planning algorithm. 
The approach combines the advantages of BTs, in terms of modularity and reactivity with the  synthesis capability of automated planning. The reactivity enables the system to both skip actions that were executed by external agents, and repeat actions that were undone by external agents. The modularity enables the extension of BTs to add new actions, when previously satisfied conditions
are violated by external agents.
Finally,  the approach was illustrated in a dynamic and challenging scenario.


\bibliographystyle{IEEEtran}
\bibliography{actionPlanningRefs,behaviorTreeRefs}
\end{document}